# DRSLF: Double Regularized Second-Order Low-Rank Representation for Web Service QoS Prediction

Hao Wu
School of Information and Software Engineering
University of Electronic Science and Technology of China
Chengdu, China
wuhaoxy@yeah.net

Jialiang Wang
College of Computer and Information Science
Southwest University
Chongqing, China
goallow@163.com

*Abstract*—Quality-of-Service (QoS) data plays a crucial role in cloud service selection. Since users cannot access all services, QoS can be represented by a high-dimensional and incomplete (HDI) matrix. Latent factor analysis (LFA) models have been proven effective as low-rank representation techniques for addressing this issue. However, most LFA models rely on first-order optimizers and use $L_2$-norm regularization, which can lead to lower QoS prediction accuracy. To address this issue, this paper proposes a double regularized second-order latent factor (DRSLF) model with two key ideas: a) integrating $L_1$-norm and $L_2$-norm regularization terms to enhance the low-rank representation performance; b) incorporating second-order information by calculating the Hessian-vector product in each conjugate gradient step. Experimental results on two real-world response-time QoS datasets demonstrate that DRSLF has a higher low-rank representation capability than two baselines.

*Keywords—Quality-of-Service, Cloud Service, Low-Rank Representation, Latent Factor Analysis, Regularization, Second-Order Optimization*

## I. Introduction

With the rapid growth of cloud computing, a vast array of cloud services is now available online [1-20]. The non-functional performance of these services is typically measured through various Quality-of-Service (QoS) metrics, *e.g.*, response-time, throughput, and failure probability [20-35]. The QoS data plays a critical role in cloud service selection.

In real-world scenarios, QoS data is often represented as matrices, such as response-time, throughput, or failure probability matrices. However, it is impractical for each user to access all services due to the large number of users and services. As a result, the QoS matrix is typically high-dimensional and incomplete (HDI) [36-50]. Although the QoS matrix is HDI, *i.e.*, with existing massive missing data, it contains abundant knowledge and interaction patterns. Extracting the knowledge from the sparse QoS data becomes a critical issue.

The latent factor analysis (LFA) model, a kind of low-rank representation method, is widely used in predicting QoS missing data. It supposes that the users' and services' features can be represented in a low-rank latent factor space, and the QoS data can be approximated by the inner product of the user latent factor vector and the service latent factor vector [51-60].

Currently, the most LFA models adopt the $L_2$-norm regularization term to avoid the problem of overfitting [9-14, 57, 58] However, the QoS data contains outliers, and the $L_2$-norm regularization is sensitive to these outliers, which may reduce the low-rank representation capability of the LFA model. Moreover, most LFA models are optimized by first-order optimizers, *e.g.*, gradient descent, gradient descent with momentum, and adaptive gradient descent. However, the objective function of the LFA model is bi-linear and non-convex. The first-order LFA model only considers the gradient information of each training period, which makes it easy to fall into a stationary point in the process of optimizing the objective function of the LFA model. The second-order optimization method not only considers the current gradient information of a certain point, but also the curvature information, which can achieve higher solution accuracy. However, the second-order optimization algorithm, *e.g.*, Newton's method, needs to calculate and store the Hessian matrix and its inverse, where the cost of calculating and storing the inverse of the Hessian matrix requires the square and cube of the model parameters, respectively.

To address the above issues, we propose a low-rank representation method for web service QoS prediction, *i.e.*, a double regularized second-order latent factor analysis model, termed DRSLF, with the following two-fold idea:

1) We integrate $L_1$-norm and $L_2$-norm regularization terms in the LFA model's objective function to reduce the sensitivity of the LFA model for outliers.

2) The DRSLF model incorporates the second-order information by calculating multiple Hessian-vector in each conjugate gradient step.

The experimental results on two QoS datasets show that the proposed DRSLF model has better low-rank representation ability than the two LFA models based on first-order and second-order optimizers.

The rest of this paper is organized as follows: Section II provides the preliminaries. Section III presents the proposed DRSLF model. Section IV gives the experimental results. Section V draws the conclusion.

## II. Preliminaries

### A. Problem Statement

**Definition 1. (A QoS Matrix)** Given two entry sets, e.g., user set $U$ and service set $S$, $\mathbf{Q} \in \mathbb{R}^{|U| \times |S|}$ describes the big picture between $U$ and $S$. In detail, each element of $\mathbf{Q}$, i.e., scalar $q_{u,s}$, denotes a QoS historical value produced by user $u \in U$ and service $i \in S$. Let $K$ and $N$, organized by a list of QoS record tuples, e.g., $(u, s, q_{u,s})$, denote known and unknown sets of $\mathbf{Q}$, respectively. The $\mathbf{Q}$ is HDI if and only if the scale of $|K| \ll |M|$.

**Definition 2. (An LFA Model)** Given set $U$, $S$, and $K$, the task of a vanilla LFA model constructs a low-rank representation of $\mathbf{Q}$, i.e., $\mathbf{Q} \approx \hat{\mathbf{Q}} = \mathbf{X}_U \mathbf{X}_S^T$, where $\mathbf{X}_U \in \mathbb{R}^{|U| \times f}$ and $\mathbf{X}_S \in \mathbb{R}^{|S| \times f}$ denote the embedding matrix of $U$ and $S$, respectively. And the estimation can be obtained by embedding vectors' inner product, i.e., $\hat{q}_{u,s} = \mathbf{x}_u \mathbf{x}_s^T$, where $\mathbf{x}_u \in \mathbb{R}^f$ and $\mathbf{x}_s \in \mathbb{R}^f$ denote the $u$-th row and $s$-th row of $\mathbf{X}_U$ and $\mathbf{X}_S$, respectively.

To achieve the optimal estimation, a loss function to measure the gap between the target $\mathbf{Q}$ and the estimation $\hat{\mathbf{Q}}$ is desired. According to [9-14, 57, 58], the $L_2$-norm (a.k.a. Euclidean distance) loss function is adopted to model such a gap. The loss function of a vanilla LFA model is given as follows:

$$L(\mathbf{X}) = \frac{1}{2} \sum_{q_{u,s} \in K} \left( q_{u,s} - \mathbf{x}_u \mathbf{x}_s^T \right)^2 = \frac{1}{2} \sum_{q_{u,s} \in K} \left( q_{u,s} - \sum_{d=1}^{f} x_{u,d} x_{i,d} \right)^2, \quad (1)$$
$$s.t. \ \forall u \in U, s \in S, d \in \{1,...,f\},$$

where $\mathbf{X} = vec(\mathbf{X}_U, \mathbf{X}_S) \in \mathbb{R}^{(|U|+|S|) \times f}$ denotes the decision parameter vector, $vec()$ denotes the vectorized representation, $\mathbf{x}_u \in \mathbb{R}^f$ and $x_{u,d}$ denote the $u$-th sub-vector of $\mathbf{X}$ and $d$-th element of $\mathbf{x}_u$, respectively. The same definition as $\mathbf{x}_s$ and $x_{s,d}$.

### B. A Second-order-based Latent Factor Analysis Model

According to [59-70], a second-order latent factor model performs Taylor expansion at point $\mathbf{X} \in \mathbb{R}^{(|U|+|S|) \times f}$ and aims to minimize the following function:

$$\arg\min \mathbf{g}_L(\mathbf{X}) + \mathbf{H}_L(\mathbf{X}) \Delta \mathbf{X} \leq \tau \mathbf{1}, \quad (2)$$

where $\mathbf{g}_L(\mathbf{X}) \in \mathbb{R}^{(|U|+|S|) \times f}$ denotes the gradient vector of loss function $L(\mathbf{X})$, $\mathbf{G}_L(\mathbf{X}) \in \mathbb{R}^{((|U|+|S|) \times f) \times ((|U|+|S|) \times f)}$ denotes the Hessian matrix or its variants, e.g., Fisher matrix and Gauss-Newton matrix, $\Delta \mathbf{X} \in \mathbb{R}^{(|U|+|S|) \times f}$ denotes the incremental vector at point $\mathbf{X}$, $\mathbf{1} \in \mathbb{R}^{(|U|+|S|) \times f}$ denotes the all-one vector, $\tau$ denotes tolerance controlling the termination of second-order optimization, when $\tau=0$, Eq. (2) degenerates into the standard Newton's method.

### C. A Double Regularization-based Latent Factor Analysis Model

The double regularization-based LFA model is as follows:

$$E(\mathbf{X}) = L(\mathbf{X}) + L_{R1}(\mathbf{X}) + L_{R2}(\mathbf{X})$$
$$= \frac{1}{2} \sum_{q_{u,s} \in K} \left( q_{u,s} - \sum_{d=1}^{f} x_{u,d} x_{s,d} \right)^2 + \sum_{d=1}^{f} \left( x_{u,d}^2 + x_{s,d}^2 \right) + \sum_{d=1}^{f} \left( \sqrt{x_{u,d}^2 + \varepsilon} + \sqrt{x_{s,d}^2 + \varepsilon} \right), \quad (3)$$

where $L_{R1}(\mathbf{X})$ denotes the smooth $L_1$-norm regularization term, $L_{R2}(\mathbf{X})$ denotes the $L_2$-norm regularization term, $\varepsilon$ is a small positive constant

## III. A Double Regularized SLF Model

### A. Hidden Mapping

Due to the bi-linear nature of the LFA model's loss function, i.e., $\mathbf{x}_u \mathbf{x}_s^T$, it is difficult to perform second-order analysis directly. Based on [59-70], mapping the LFA model's bi-linear term into a hidden function can simplify such an analysis process. The mapping process is given as follows:

$$H(\mathbf{X})_{u,s} = \mathbf{x}_u \mathbf{x}_s^T = \sum_{d=1}^{f} x_{u,d} x_{s,d}, \quad (4)$$

where $H(\cdot)$ denotes the hidden function, $H(\mathbf{X})_{u,s}$ denotes the mapping value. Then, the loss function $L(\mathbf{X})$ can be reformulated as follows:

$$L(H(\mathbf{X})) = \frac{1}{2} \sum_{q_{u,s} \in K} \left( q_{u,s} - H(\mathbf{X})_{u,s} \right). \tag{5}$$

*B. Gauss-Newton Approximation*

The loss function of LFA is non-convex due to its bi-linear nature. Therefore, the Hessian matrix of the LFA model is indefinite. According to [59-70], for non-convex representation learning model, performing the Gauss-Newton approximation can obtain the semi-positive curvature matrix. The Gauss-Newton matrix of $L(H(\mathbf{X}))$ can be calculated as follows:

$$\mathbf{G}_L(H(\mathbf{X})) = \mathbf{J}_H(\mathbf{X})^\mathrm{T} \mathbf{J}_H(\mathbf{X}), \tag{6}$$

where $\mathbf{G}_L(H(\mathbf{X})) \in \mathbb{R}^{((|U|+|S|) \times f) \times ((|U|+|S|) \times f)}$ denotes the Gauss-Newton matrix of $L(H(\mathbf{X}))$, $\mathbf{J}_H(\mathbf{X}) \in \mathbb{R}^{|K| \times ((|U|+|S|) \times f)}$ denotes the Jacobian matrix of $H(\mathbf{X})$.

*C. Hessian-Vector Product*

For the latent factor model, applying second-order methods, such as the Newton method, requires substantial computational and storage resources, *i.e.*, $O(((|U|+|S|) \times f)^2)$ to manipulate curvature matrix and $O(((|U|+|S|) \times f)^3)$ to compute its inverse. However, we do not need to manipulate the curvature matrix and its inverse directly. Performing conjugate gradient and calculating Hessian-vector in each conjugate gradient step can acquire the curvature matrix with an acceptable cost [59-70]. The details of computing the Hessian-vector in each step can be derived as follows:

$$\begin{aligned}\omega_L(H(\mathbf{X})) &= \mathbf{G}_L(H(\mathbf{X}))\mathbf{v} \\ &= \mathbf{J}_H(\mathbf{X})^\mathrm{T} \mathbf{J}_H(\mathbf{X})\mathbf{v}, \end{aligned} \tag{7}$$

where $\omega_L(H(\mathbf{X})) \in \mathbb{R}^{(|U|+|S|) \times f}$ denotes Hessian-vector product, and $\mathbf{v} \in \mathbb{R}^{(|U|+|S|) \times f}$ denotes the conjugate direction in each conjugate gradient iteration. The Jacobian matrix $\mathbf{J}_H(\mathbf{X})$ can be derived as follows:

$$\mathbf{J}_\sigma(\mathbf{X}) = \left( \frac{\partial}{\partial \mathbf{X}} H(\mathbf{X})_{u,s} \Big|_{(u,s) \in K} \right). \tag{8}$$

Then, the Jacobian-vector, *i.e.*, $\mathbf{J}_H(\mathbf{X})\mathbf{v} \in \mathbb{R}^{|K|}$ can be calculated as follows:

$$\begin{aligned}\mathbf{J}_H(\mathbf{X})\mathbf{v} &= R\left\{ H(\mathbf{X})_{u,s} \Big|_{(u,s) \in K} \right\} \\ &= \left( \sum_{d=1}^{f} \left( v_{u,d} x_{s,d} + x_{u,d} v_{s,d} \right) \Big|_{(u,s) \in K} \right), \end{aligned} \tag{9}$$

where $R\{\cdot\}$ denotes the *R*-operator [59-70].

Next, we can obtain the Hessian-vector by calculating the vector-Jacobian product, *i.e.*, $\omega_L(H(\mathbf{X})) = (\mathbf{J}_H(\mathbf{X})\mathbf{v})^\mathrm{T} \mathbf{J}_H(\mathbf{X})$. Combining *Eq.* (8) and *Eq.* (9), we have:

$$\begin{cases} \forall u \in U, d = 1 \sim f: \\ \omega_L(\sigma(\mathbf{X}))_{u,d} = \sum_{i \in K_u} \left( x_{s,d} \left( \sum_{d=1}^{f} (v_{u,d} x_{s,d} + x_{u,d} v_{s,d}) \right) \right) \\ \forall s \in S, d = 1 \sim f: \\ \omega_L(\sigma(\mathbf{X}))_{s,d} = \sum_{s \in K_s} \left( x_{u,d} \left( \sum_{d=1}^{f} (v_{u,d} x_{s,d} + x_{u,d} v_{s,d}) \right) \right) \end{cases} \tag{10}$$

where $K_u$ and $K_s$ denote the user *u*'s known set and service *s*' known set, respectively.

*D. Double Regularization Term Incorporation*

Recall $E(\mathbf{X})$, the objective function of the double regularized latent factor model, whose regularization is the linear combination of $L_1$-norm regularization and $L_2$-norm regularization. Hence, the Hessian-vector product of $E(\mathbf{X})$'s regularization can be derived as follows:

$$\omega_R(\mathbf{X}) = \omega_{R1}(\mathbf{X}) + \omega_{R1}(\mathbf{X}), \tag{11}$$

where $\omega_{R1}(\mathbf{X}) \in \mathbb{R}^{(|U|+|S|) \times f}$ denotes the $L_1$-norm regularization term's Hessian-vector, and $\omega_{R2}(\mathbf{X}) \in \mathbb{R}^{(|U|+|S|) \times f}$ denotes the $L_2$-norm regularization term's Hessian-vector product. In detail, the $L_1$-norm regularization term's Hessian-vector $\omega_{R1}(\mathbf{X})\mathbf{v}$ in each conjugate gradient iteration can be computed as follows:

$$\omega_{R1}(\mathbf{X}) \Rightarrow \begin{cases} \forall u \in \mathbf{U}, d = 1 \sim f: \\ \omega_{R1}(\mathbf{X})_{u,d} = \dfrac{\lambda_{R1} v_{u,d} |\mathbf{K}_u| \varepsilon}{(x_{u,d}^2 + \varepsilon)\sqrt{x_{u,d}^2 + \varepsilon}}, \\ \forall s \in \mathbf{S}, d = 1 \sim f: \\ \omega_{R1}(\mathbf{X})_{s,d} = \dfrac{\lambda_{R1} v_{s,d} |\mathbf{K}_s| \varepsilon}{(x_{u,d}^2 + \varepsilon)\sqrt{x_{s,d}^2 + \varepsilon}}, \end{cases} \quad (12)$$

where $\lambda_{R1}$ and $\varepsilon$ are hyperparameters, $|\mathbf{K}_u|$ and $|\mathbf{K}_s|$ denote the volume of set $\mathbf{K}_u$ and $\mathbf{K}_s$, respectively. And the $L_2$-norm regularization term's Hessian-vector product $\omega_{R2}(\mathbf{X})v$ can be computed as follows:

$$\omega_{R2}(\mathbf{X}) \Rightarrow \begin{cases} \forall u \in \mathbf{U}, d = 1 \sim f: \\ \omega_{R2}(\mathbf{X})_{u,d} = \lambda v_{u,d} |\mathbf{K}_u|, \\ \forall s \in \mathbf{S}, d = 1 \sim f: \\ \omega_{R2}(\mathbf{X})_{i,d} = \lambda v_{s,d} |\mathbf{K}_s|, \end{cases} \quad (13)$$

Then, combining *Eq.* (10), *Eq.* (12), and *Eq.* (13), we can obtain the double regularized LFA model's objective function's Hessian-vector product as follows:

$$\omega_E(\mathbf{X}) \approx \mathbf{G}_L(\mathbf{X})v + \mathbf{G}_{R1}(\mathbf{X})v + \mathbf{G}_{R2}(\mathbf{X})v \quad (14)$$

where $\omega_E(\mathbf{X}) \in \mathbb{R}^{(|U|+|S|) \times f}$ denotes the Hessian-vector product of *Eq.* (3). Its single element dependent form can be expanded as:

$$\omega_E(\mathbf{X}) \Leftarrow \begin{cases} \forall u \in \mathbf{U}, d = 1 \sim f: \\ \omega_E(\mathbf{X})_{u,d} = \sum_{i \in \mathbf{K}_u} \left( x_{i,d} \left( \sum_{d=1}^{f} (v_{u,d} x_{s,d} + x_{u,d} v_{s,d}) \right) \right) \\ \qquad + \dfrac{\lambda_{R1} v_{u,d} |\mathbf{K}_u| \varepsilon}{(x_{u,d}^2 + \varepsilon)\sqrt{x_{u,d}^2 + \varepsilon}} + \lambda v_{u,d} |\mathbf{K}_u|, \\ \forall s \in = \mathbf{S}, d = 1 \sim f: \\ \omega_E(\mathbf{X})_{s,d} = \sum_{u \in \mathbf{K}_i} \left( x_{u,d} \left( \sum_{d=1}^{f} (v_{u,d} x_{s,d} + x_{u,d} v_{isd}) \right) \right) \\ \qquad + \dfrac{\lambda_{R1} v_{s,d} |\mathbf{K}_s| \varepsilon}{(x_{s,d}^2 + \varepsilon)\sqrt{x_{s,d}^2 + \varepsilon}} + \lambda v_{i,d} |\mathbf{K}_i|. \end{cases} \quad (15)$$

### E. Damping Term Incorporation

Note that applying the Gauss-Newton approximation for the objective function of the double regularization LFA model can obtain a semi-positive curvature matrix. Incorporating a sufficient damping term into the Gauss-Newton matrix, regarded as a regularization term for curvature matrix balancing, first-order and second-order approximation, can keep it positive [59-70], and avoid obtaining incorrect update directions through conjugate gradient. The details of adding a damping term to the curvature matrix are given as follows:

$$\omega_E(\mathbf{X}) \Leftarrow \begin{cases} \forall u \in \mathbf{U}, d = 1 \sim f: \\ \omega_E(\mathbf{X})_{u,d} = \sum_{i \in \mathbf{K}_u} \left( x_{i,d} \left( \sum_{d=1}^{f} (v_{u,d} x_{s,d} + x_{u,d} v_{s,d}) \right) \right) \\ \qquad + \dfrac{\lambda_{R1} v_{u,d} |\mathbf{K}_u| \varepsilon}{(x_{u,d}^2 + \varepsilon)\sqrt{x_{u,d}^2 + \varepsilon}} \\ \qquad + \lambda v_{u,d} |\mathbf{K}_u| \\ \qquad + \gamma v_{u,d}, \\ \forall i \in \mathbf{I}, d = 1 \sim f: \\ \omega_E(\mathbf{X})_{s,d} = \sum_{u \in \mathbf{K}_i} \left( x_{u,d} \left( \sum_{d=1}^{f} (v_{u,d} x_{s,d} + x_{u,d} v_{s,d}) \right) \right) \\ \qquad + \dfrac{\lambda_{R1} v_{s,d} |\mathbf{K}_s| \varepsilon}{(x_{u,d}^2 + \varepsilon)\sqrt{x_{s,d}^2 + \varepsilon}} \\ \qquad + \lambda v_{s,d} |\mathbf{K}_s| \\ \qquad + \gamma v_{s,d} \end{cases} \quad (16)$$

where $\mathbf{I} \in \mathbb{R}^{((|U|+|S|) \times f) \times ((|U|+|S|) \times f)}$ denotes the identity matrix, and $\gamma$ denotes the damping term.

*F. Update Rule*

After multiple conjugate gradient iterations, the update direction at the *t*-th epoch can be derived as follows:

$$\begin{cases} \Delta \mathbf{X}^t \stackrel{Conjugate}{\Leftarrow} \left(\mathbf{G}_E(\mathbf{X})^t + \gamma \mathbf{I}\right) \Delta \mathbf{X}^t + \mathbf{g}_E(\mathbf{X}) \leq \tau \mathbf{1}, \\ \mathbf{X}^{t+1} = \mathbf{X}^t + \Delta \mathbf{X}^t, \end{cases} \quad (17)$$

IV. EXPERIMENTS

In this section, all experiment details are given.

*A. General Settings*

**Datasets.** The response-time dataset, an open-source QoS data collected by WS-Dream, is adopted in this section. The scale of response-time QoS matrix is 339 by 5825, which contains 1,974,675 elements. The response-time dataset is divided into a training set, a validation set, and a test set. The details of the testing cases of this dataset are shown in Table I.

TABLE I. INVOLVED DATASETS

| No. | Train | Test | Validation | Density |
|---|---|---|---|---|
| D1 | 10% | 45% | 45% | 10% |
| D2 | 20% | 40% | 40% | 20% |

**Evaluation metric.** We adopt the root mean square error (RMSE) to evaluate the prediction performance of the proposed method and other competitors. The lower the RMSE value, the better the prediction performance for missing elements in the HDI matrix [71-75].

The Root Mean Square Error (RMSE) is adopted to evaluate the low-rank representation performance.

$$RMSE = \sqrt{\frac{\sum_{q_{u,s} \in \Omega}\left(r_{u,i} - \sum_{d=1}^{f} x_{u,d} x_{s,d}\right)^2}{|\Omega|}} \quad (18)$$

where $\Omega$ denotes the evaluation set.

**Competitors.** The proposed DRSLF model is compared to the baseline optimizer-based LFA models as follows:

*1)* **An SGDM-based LFA model (M1)** [66]**:** The vanilla SGD is a default optimizer for any representation learning model, including the vanilla LFA model.
*2)* **An SLF model (M2)** [64]**:** A second-order LFA model via the Hessian-vector product method.
*3)* **A DRSLF model (M3):** The proposed PSLF model in this paper.

**Setting Strategy.** The detailed information regarding all benchmark datasets and the settings of the compared models is as follows:

*1) Environment Configuration:* All experiments are conducted on a laptop with Windows 11 Pro, Intel Core 13905H 5.4 GHz, and 32 GB RAM. Moreover, all tested models are running on OpenJDK 11 LTS.

*2) Hyperparameter Optimization:* We fixed the dimension of *f* as 20, and the elements of the latent matrices, *i.e.*, $\mathbf{X}_U$ and $\mathbf{X}_S$, are sampled from $U(0,0.04)$. For M1 and M2, their hyperparameters are fine-tuned according to their official guidelines [64, 66]. For M5, the search range of $\lambda_{R1}$ is as [0.0,0.01,…,0.1], the search range of $\lambda_{R2}$ is $[10^{-7},10^{-6},…,10^{-3}]$, the search range of damping term $\gamma$ is set at [20,40,…,300], the tolearance $\tau$ is fixed as 10.

*3) Terminate Condition:* The maximum training epoch is set as 500. We adopt the early stop termination strategy in this section, and the maximum early stop epoch is set as 10.

*B. Comparison Results*

The comparison results are summarized in Table IV. From Table IV, we can make the following observations:

**(1) The curvature information can enhance the prediction accuracy of the LFA model's low-rank representation.** On the D1 dataset, the average RMSE for M1, the first-order stochastic gradient-based LFA model, is 1.37776. For the second-order LFA models, M2 and M3, their average RMSE values are 1.37762 and 1.37029, which are 0.01% and 0.54% lower than that of M1, respectively. On the D2 dataset, the average RMSE values for M1, M2, and M3 are 1.29838, 1.29363, and 1.28743, respectively. In other words, for M2 and M3, their average RMSE values are 0.37% and 0.84% lower than that of M1 on D2, respectively.

**(2) Incorporating the $L_1$-norm regularization term can further enhance the low-rank representation ability.** On the D1 dataset, the average RMSE for M3 is 1.37029, while for M2, it is 1.37762, indicating a 0.53% improvement for M3. On the D2 dataset, the average RMSE for M3 is 1.28743, and for M2, it is 1.29363, showing a 0.48% improvement for M3.

TABLE II. PERFORMANCE BENCHMARK ON D1-D5

| Datasets | Model | RMSE | Epoch |
|---|---|---|---|
| D1 | M1 | 1.37776 | **30** |
| | M2 | 1.37762 | 38 |
| | M3 | **1.37029** | 39 |
| D2 | M1 | 1.29838 | 37 |
| | M2 | 1.29363 | 41 |
| | M3 | **1.28743** | **36** |

## V. CONCLUSION

In this paper, we propose a low-rank representation method, termed by double-regularized second-order latent factor model, for predicting QoS data. The model enhances low-rank representation capabilities by incorporating both $L_1$-norm and $L_2$-norm regularization terms. Experimental results demonstrate that the proposed model outperforms two baseline latent factor models in terms of low-rank representation capability. In future work, we plan to extend this method to tensor representation learning tasks.